\documentclass[default]{sn-jnl} 

\usepackage{graphicx}%
\usepackage{multirow}%
\usepackage{amsmath,amssymb,amsfonts}%
\usepackage{amsthm}%
\usepackage{mathrsfs}%
\usepackage[title]{appendix}%
\usepackage{xcolor}%
\usepackage{textcomp}%
\usepackage{manyfoot}%
\usepackage{booktabs}%
\usepackage{algorithm}%
\usepackage{algorithmicx}%
\usepackage{algpseudocode}%
\usepackage{listings}%
\usepackage{subcaption}

\begin{document}

\title[Article Title]{KinD-LCE: Curve Estimation and Retinex Fusion on Low-Light Image}


\author*[1,2]{\fnm{Xiaochun} \sur{Lei}}\email{glleixiaochun@qq.com}

\author[1,2]{\fnm{Weiliang} \sur{Mai}}\email{ruomengawa@gmail.com}

\author[1,2]{\fnm{Junlin} \sur{Xie}}\email{junlinxie@email.cn}

\author[1,2]{\fnm{He} \sur{Liu}}\email{2973720373@qq.com}

\author[1,2]{\fnm{Zetao} \sur{Jiang}}\email{zetaojiang@guet.edu.cn}

\author[1,2]{\fnm{Zhaoting} \sur{Gong}}\email{gavin@gong.host}

\author[1,2]{\fnm{Chang} \sur{Lu}}\email{me@keter.top}

\author[1,2]{\fnm{Linjun} \sur{Lu}}\email{me@zerorains.top}

\affil*[1]{\orgdiv{School of Computer Science and Information Security}, \orgname{Guilin University of Electronic Technology}, \orgaddress{\street{Jinji Road}, \city{Guilin}, \postcode{541004}, \state{Guangxi}, \country{China}}}

\affil[2]{\orgdiv{Guangxi Key Laboratory of Image and Graphic Intelligent Processing}, \orgname{Guilin University of Electronic Technology}, \orgaddress{\street{Jinji Road}, \city{Guilin}, \postcode{541004}, \state{Guangxi}, \country{China}}}


\abstract{Low-light images often suffer from noise and color distortion. Object detection, semantic segmentation, instance segmentation, and other tasks are challenging when working with low-light images because of image noise and chromatic aberration. We also found that the conventional Retinex theory loses information in adjusting the image for low-light tasks. In response to the aforementioned problem, this paper proposes an algorithm for low illumination enhancement. The proposed method, KinD-LCE, uses a light curve estimation module to enhance the illumination map in the Retinex decomposed image, improving the overall image brightness. An illumination map and reflection map fusion module were also proposed to restore the image details and reduce detail loss. Additionally, a TV(total variation) loss function was applied to eliminate noise. Our method was trained on the GladNet dataset, known for its diverse collection of low-light images, tested against the Low-Light dataset, and evaluated using the ExDark dataset for downstream tasks, demonstrating competitive performance with a PSNR of 19.7216 and SSIM of 0.8213.}

\maketitle
\section{Introduction}
\label{sec1}

Low-light environments are a great challenge in the field of computer vision, significantly impacting the quality and utility of captured images due to issues such as noise, blurring, and color distortion. Traditional computer vision algorithms, while proficient in handling images captured under standard lighting conditions, often falter when applied to low-light images, leading to diminished image quality and, consequently, reduced efficiency in industrial production and other practical applications.

One prevalent approach to mitigating the impact of low-light conditions on image quality involves leveraging the Retinex theory, which, despite its effectiveness, assumes a smooth illumination map and thus can introduce issues such as edge blurring and increased noise. Furthermore, while existing algorithms, such as histogram equalization, have been utilized to enhance low-light images, they come with their own set of limitations, including unnatural equalization effects, a propensity for over-enhancement, and suboptimal computational efficiency.

\begin{figure}[h]%
\centering
\includegraphics[width=0.9\textwidth]{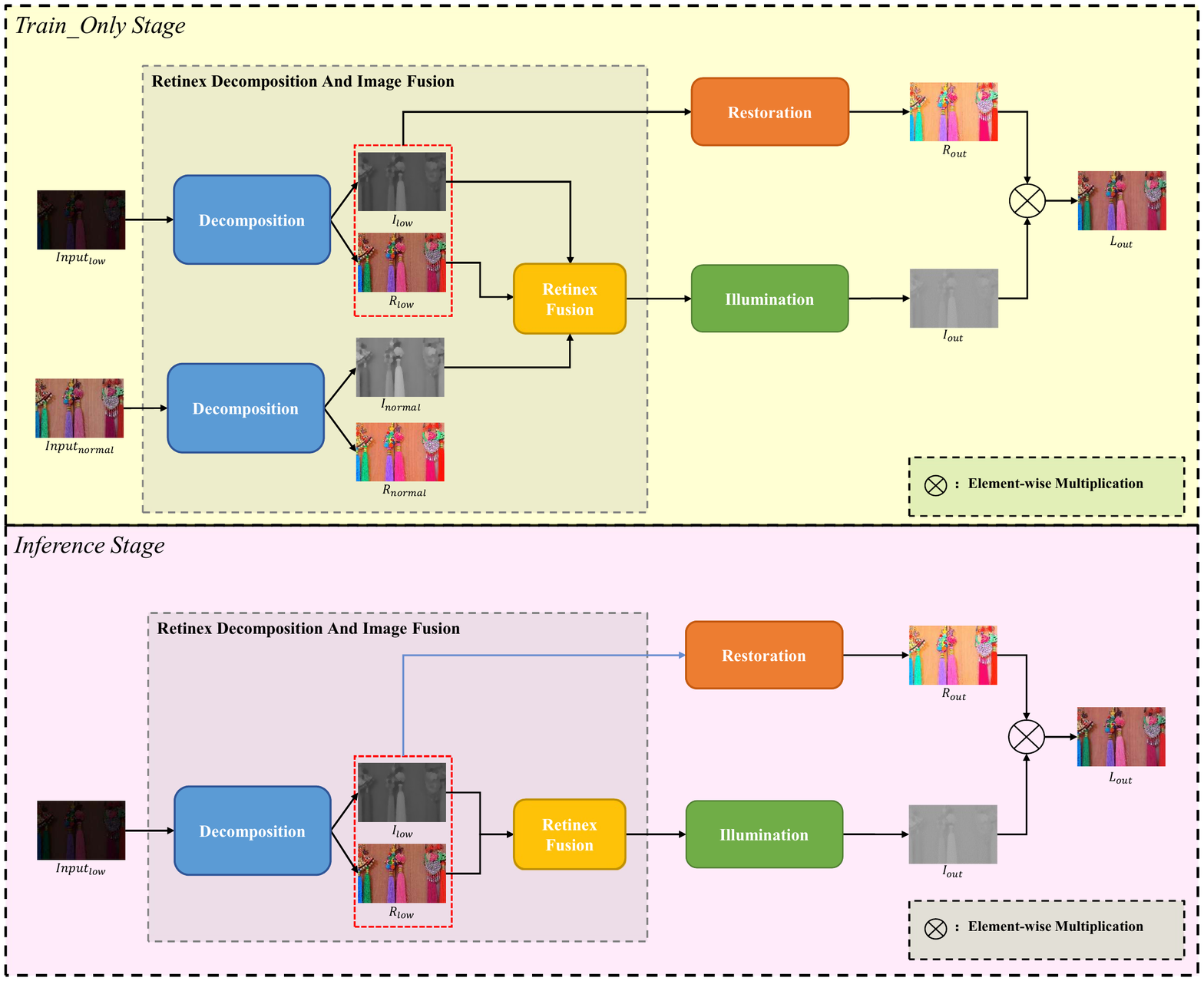}
\caption{Schematic diagram of KinD-LCE network structure}\label{fig4.1}
\end{figure}

Concerning the aforementioned situation and issues, we proposed a Retinex Fusion module based on the U-shaped network structure, which extends the Retinex theory and fuses the features of the reflectance map with the illumination map to solve the edge-blurring problem in the illumination map, followed by the light curve estimation module to adjust the image brightness distribution by the trainable parameter $\alpha$ to reduce the noise generated during upsampling and downsampling, to reduce the edge detail loss, and finally, to suppress the generation of noise with the total variation in the restoration network loss function.

The main contributions of this study are as follows.
\begin{enumerate}
    \item The feature information contained in the reflectance map of the image is fused with the illumination map of the image to optimize the image quality, and the image brightness is enhanced using the reflectance map to calculate the illumination coefficients.
    \item The brightness of the image is improved with the Light Curve Estimation (LCE) module, which brings out more detail in the image.
    \item The TV loss function is included in the restoration network to further eliminate the noise, reduce the parts with significant variation, and smooth the image edges.
\end{enumerate}

The rest of the paper is organized as follows. We describe our proposed approach in Sec.\ref{sec4}. We demonstrate our methods in Sec.\ref{sec5}. We discuss relevant work in Sec.\ref{sec2} and conclude in Sec.\ref{sec6}.

\section{Related Works}\label{sec2}

The analysis and solution presented in this section are motivated by the increasing significance of handling low-light images in computer vision. We begin by examining our previous work and then introduce our novel approach.

\subsection{Low-Light Enhancement}\label{subsec21}

The primary task of low-light enhancement is to increase the brightness of the image to improve the recognition accuracy of computer vision tasks, such as image classification, target detection, and image segmentation, and to reduce the effect of environmental factors on recognition accuracy.

Histogram equalization \cite{pizer1987adaptive} is the most widely used low-illumination enhancement algorithm in early research. It uses a cumulative distribution function to normalize the pixels of an image to the entire grey level, by which the histogram of the output image is constrained to enhance the overall image contrast. Simultaneously, it is possible to calculate the parameters for block-wise histogram equalization, thereby avoiding the excessive enhancement that results from global equalization. This method is simple to implement, suitable for various images, and produces better results than many conventional processing methods. The technology has certain limitations in low-light enhancement. Specifically, in low-light images, the limited information in the image and the presence of noise can lead to issues, such as over-enhancement, excessive denoising, and distortion, when using global histogram equalization and other global histogram adjustment algorithms. Additionally, these methods may increase the noise level of the image, decreasing the image quality. Therefore, we should comprehensively pay more attention to local control to enhance the effectiveness of low-light enhancement tasks. 

The Retinex model \cite{land1971lightness} and its multi-scale version \cite{zhou2020multiscale, jobson1997multiscale} decompose the image into illumination and reflection maps and process the two maps separately before fusing them to obtain the recovered image. 
The model can process images based on the human eye's perception of color and simulate changes in a scene's lighting conditions, allowing for a better visual experience in low light or other adverse environments. 
However, it is difficult to achieve better robustness using only hand-designed a priori knowledge. 
The grayscale values of images are very close to each other in low-light conditions, making it difficult to distinguish them during multi-scale decomposition, resulting in information loss and the generation of artifacts, and it is deeply affected by noise. 
These factors can lead to an inability to recover details, colors, and other features in the image. 
Therefore, it is necessary to dynamically adjust prior knowledge according to the current environment and preserve more loss information in the multi-scale decomposition process [19,35].

Inspired by Photoshop-like image editing software, researchers \cite{guo2020zero} have recently attempted to design a curve-based adjustment method that adaptively adjusts an image's light curve. 
A low-light image is mapped automatically to a light curve, and its brightness is adjusted according to the light curve. There are three goals for such a curve: 1) The pixel values of the image to be enhanced should be normalized to $[0, 1]$ to avoid gradient explosion because of large loss values; 2) The curve should be monotonic to facilitate maintaining the variance between adjacent pixels; 3) Backpropagation to compute the gradient should be as derivable as possible. In practical applications, this method effectively solves the gamma correction problem and achieves high fidelity. However, as a curve estimation-based method, the instability and limitations of the results may lead to the destruction of the original image features, affecting the original image's structure. 

So another group of researchers \cite{rahman2020efficient} came up with an alternative method of automatic image enhancement. They first used a convolutional neural network to determine its category. Then, images are converted into photonegative form to obtain an initial transmission map using a bright channel prior. Next, L1-norm regularization is adopted to refine scene transmission. Besides, environmental light is estimated based on an effective filter. Finally, the image degradation model is applied to achieve enhanced results. The method uses the prior knowledge of the image to recover the features of the original image, so as to keep the structure of the original image unchanged as much as possible, but the restoration effect may be poor because the information gap between the original image and the normal light image is too large.

Meanwhile, several studies have focused on using unsupervised, zero-shot, and few-shot learning methods to achieve low-light enhancement. 
Guo et al. \cite{guo2020zero, li2021learning} recently proposed the Zero-DCE network and Zero-DCE++ network that can be trained without training data pairs with a non-referenced loss function. 
Using a zero-reference strategy reduces the burden of data preparation and increases the training speed. 
However, because of this strategy, the network may show poor enhancement results for low-light images with high noise levels.

Zhang et al. \cite{zhang2019kindling} proposed a multi-stage low-light enhancement model based on the Retinex theory. 
The decomposition and fusion of the Retinex method were based on a neural network approach; the method was divided into three main stages. 
First, the illumination and reflection maps are segmented by Encoder-Decoder with a U-shaped structure, then the reflection map is restored by the restoration network, and the illumination map is enhanced by the enhancement network. 
Finally, the enhanced and restored images are fused based on the problems of the previously mentioned Retinex theory \cite{land1971lightness}. 
Blurred edges lead to problems, such as loss of edge details and increased noise. The KinD-Plus \cite{zhang2021beyond} network was also designed based on Retinex theory, which will have the same problems. 
Besides, Jiang et al. \cite{jiang2021enlightengan} proposed an unsupervised method to learn the correspondence between low-light and normal lighting with normally lit images without using low-light images as reference objects. 
However, this training method affects the quality of images by introducing noise and loss of edge details. Until now, Liu et al. \cite{liu2021retinex} and Xie et al. 
\cite{xie2020semantically} have proposed using semantic information to guide the reconstruction of Retinex images to eliminate noise. Researchers \cite{rahman2021structure, guan2021dual} using the dual-tree complex wavelet transform (DT-CWT) that is based on wavelets. 
The RGB space is reconstructed through V channel conversion in HSV so that the image is white-balanced to eliminate color bias.
By contrast, it is more advantageous for us to enhance the Illumination stage directly to eliminate image enhancement. 
In conclusion, regardless of whether unsupervised or supervised methods are used for enhancement, whether using contrastive images or semantic information for guidance, each approach has advantages and disadvantages.

\subsection{Image Fusion}\label{subsec22}

Most neural networks use image fusion to enhance the features extracted using the backbone network, usually fusing the features extracted at different scales or categories in the middle of the neural network to make the extracted features more effective \cite{liu2016image, zhou2020multiscale, yan2016network, yan2019pecs}[20,21,23,34]. The process of downsampling the neural network causes a loss of information, resulting in the loss of details after resampling because of insufficient information. 

In the process of low-light enhancement, the introduction of image fusion and feature fusion can compensate for the loss of details caused by the downsampling operation and preserve the details of the original image while enhancing the image's brightness. Ying et al. \cite{ying2017bio, ying2017new} argued many similarities between cameras and the human eye because the brain fuses information from different luminance ranges after the eye has acquired the image, resulting in a high dynamic range image with complete detail retention. Based on this conjecture, they proposed a method to obtain high dynamic range images by transforming the luminance of low-light images and then fusing the images of different luminance ranges.

However, because of the requirement for exposure fusion, there are specific requirements for the brightness differences between adjacent images, which may not apply to all image scenes. Moreover, compared with other low-light enhancement algorithms, the enhancement effect of this algorithm may not be natural enough, leading to excessive enhancement and loss of details in the image.

\section{Low-Light Enhancement Problems}\label{sec3}

\subsection{Basic Information}\label{subsec31}
In this paper, an image is defined as a matrix with pixel values ranging from 0 to 255, representing the degree of lightness or darkness of the image. The brightness of an image is influenced by multiple pixels and can be adjusted using factors, such as contrast and saturation. Low-light images lack scene information due to various factors, such as the environment, making it difficult for the human eye to perceive details in the image. Based on the Retinex model, the original image is decomposed into two components: the reflection map, which represents the essential properties of the image (excluding brightness) and contains edge details and color information, and the illumination map, which captures the overall scene outline and luminance distribution. And process the reflected and illuminance images to generate pictures with normal brightness.

\subsection{Problem Description}\label{subsec32}
The optimization objective of low-light image enhancement is to input two sets of images (low-light image, normal-light image) using a neural network algorithm to learn the mapping relationship from the low-light image to the normal-light image, which can be expressed by the following equation:

\begin{equation}
    \begin{aligned}
    I^{\prime}=F(I_{low},I_{normal}), \label{eq31}
    \end{aligned}
\end{equation}

where $I^\prime$ denotes the image after enhancement, $F\left(\cdot\right)$ denotes the corresponding low-light enhancement neural network, $I_{low}$ denotes the low-light image, and $I_{normal}$ denotes the normal-light image. there are various representations of $F\left(\cdot\right)$ to achieve the enhancement of low-light images, such as the histogram equalization method based on the traditional histogram equalization method HE of digital image processing algorithms \cite{pizer1987adaptive}, methods based on Retinex model \cite{zhang2019kindling, gharbi2017deep, li2021low, jiang2021enlightengan, guo2016lime, lore2017llnet}, etc. In this paper, the main recovery using Retinex model is divided into three main stages, which are image decomposition (Decomposition), illumination map enhancement (Illumination), reflection map recovery (Reflectance), and restoration of the image. In the Decomposition stage, the input image is decomposed into Illumination and Reflectance maps for subsequent processing,

\begin{equation}
    \begin{aligned}
        I,R=F_{decom}(I_{in}), \label{eq32}
    \end{aligned}
\end{equation}

where $I$ denotes the illumination map, which reflects the different brightness details of the image in different areas, $R$ denotes the reflection map, which reflects the contour details of different parts of the image after removing the brightness representation. $F_{decom}\left(\cdot\right)$ denotes the decomposition network, and $I_{in}$ denotes the input image. The input image is decomposed into an illumination map and reflection map by Decomposition, and the illumination map and reflection map are processed separately. The decomposition process requires the use of loss function optimization, which is generally used to optimize the MSE loss function and derive the minimum value of the loss function. The formula is expressed as follows.

\begin{equation}
    \begin{aligned}
        L_{mse}=\left\|\hat{I}-I_p\right\|_2^2, \label{eq33}
    \end{aligned}
\end{equation}

where $\hat{I}$ denotes the normal-light image, $I_p$ denotes the predicted image. Assuming that the parameters of the $k$-th layer are $W^{\left(k\right)}$ and $b^{\left(k\right)}$, the final optimization objective is

\begin{equation}
    \begin{aligned}
        Loss=min\left(0,\frac{\partial L_{mse}}{\partial W^{(k)}}\right), \label{eq34}
    \end{aligned}
\end{equation}

where $W^{\left(k\right)}$ denotes the weight of model, and $b^{\left(k\right)}$ denotes the bias of model. 

For the Illumination stage, the illumination maps decomposed from low-light images and normal illumination are input separately, and the enhanced illumination maps are obtained after network optimization enhancement to learn the luminance mapping relationship from low-light illumination maps to normal illumination maps. The formula is expressed as,

\begin{equation}
    \begin{aligned}
        I_{illum}=F_{illum}(I_{low},I_{normal}), \label{eq35}
    \end{aligned}
\end{equation}

where $I_{illum}$ denotes the enhanced illumination map,  $I_{low}$ denotes the illumination map obtained by decomposing the low-illumination image, and $I_{normal}$ denotes the illumination map obtained by decomposing the normal-illumination image, $F_{illum}\left(\cdot\right)$ denotes the network of the enhanced illumination map, which will accept illuminance maps under low light conditions and illuminance maps under normal light conditions to produce enhanced illuminance maps under low light conditions. The loss function is also used for optimization.

At the same time, the reflectance map obtained from the low-light image and normal light decomposition are input in the Reflectance stage respectively, and the information of the low-light reflectance map, including details and edges, is recovered by the neural network. The formula is expressed as,

\begin{equation}
    \begin{aligned}
        R_{reflect}=F_{reflect}(R_{low},R_{normal}), \label{eq36}
    \end{aligned}
\end{equation}

where $R_{reflect}$ denotes the recovered reflectance map, $R_{low}$ denotes the reflectance map obtained by decomposing the low-light image, and $R_{normal}$ denotes the reflectance map obtained by decomposing the normal-light image, $F_{reflect}\left(\cdot\right)$ denotes the network to recover the reflectance map, which will accept reflection maps under low light conditions and reflection maps under normal light conditions to produce enhanced reflection maps under low light conditions. The loss function is also used for optimization. Finally, the outputs of the Illumination and Reflectance stages are fused to obtain.

\begin{equation}
    \begin{aligned}
        I_{out}=I_{illum}\cdot R_{reflect}, \label{eq38}
    \end{aligned}
\end{equation}

Low-light image enhancement optimizes the brightness of the image by decomposing the image into an illumination map and reflection map to achieve the decoupling effect, brightness enhancement of the illumination map to achieve the recovery of the brightness of the original image, and recovery of the reflection map to avoid the introduction of excess noise.

\section{Methodology}\label{sec4}

\subsection{Network Architecture}\label{subsec41}

In this section, we present our proposed solutions to the aforementioned problems. Our approach optimized the enhancement effect while minimizing the effect on image quality. 
For the first time, we combined the Retinex image fusion module and the light curve adjustment module to solve the image restoration problem caused by incomplete decomposition in Retinex theory. The structure of the Retinex image fusion module is described in Sec.\ref{subsec43}, whereas the structure of the light adjustment curve module is discussed in Sec.\ref{subsec44}. 
To enhance image quality, we fused the illumination and reflectance maps and calculated the illumination coefficients based on the reflectance maps to suppress noise generation. We then used the TV loss function for noise removal and the illumination adjustment curve module to mitigate the effect of noise removal on luminance. 

\begin{figure}[h]%
\centering
\includegraphics[width=0.9\textwidth]{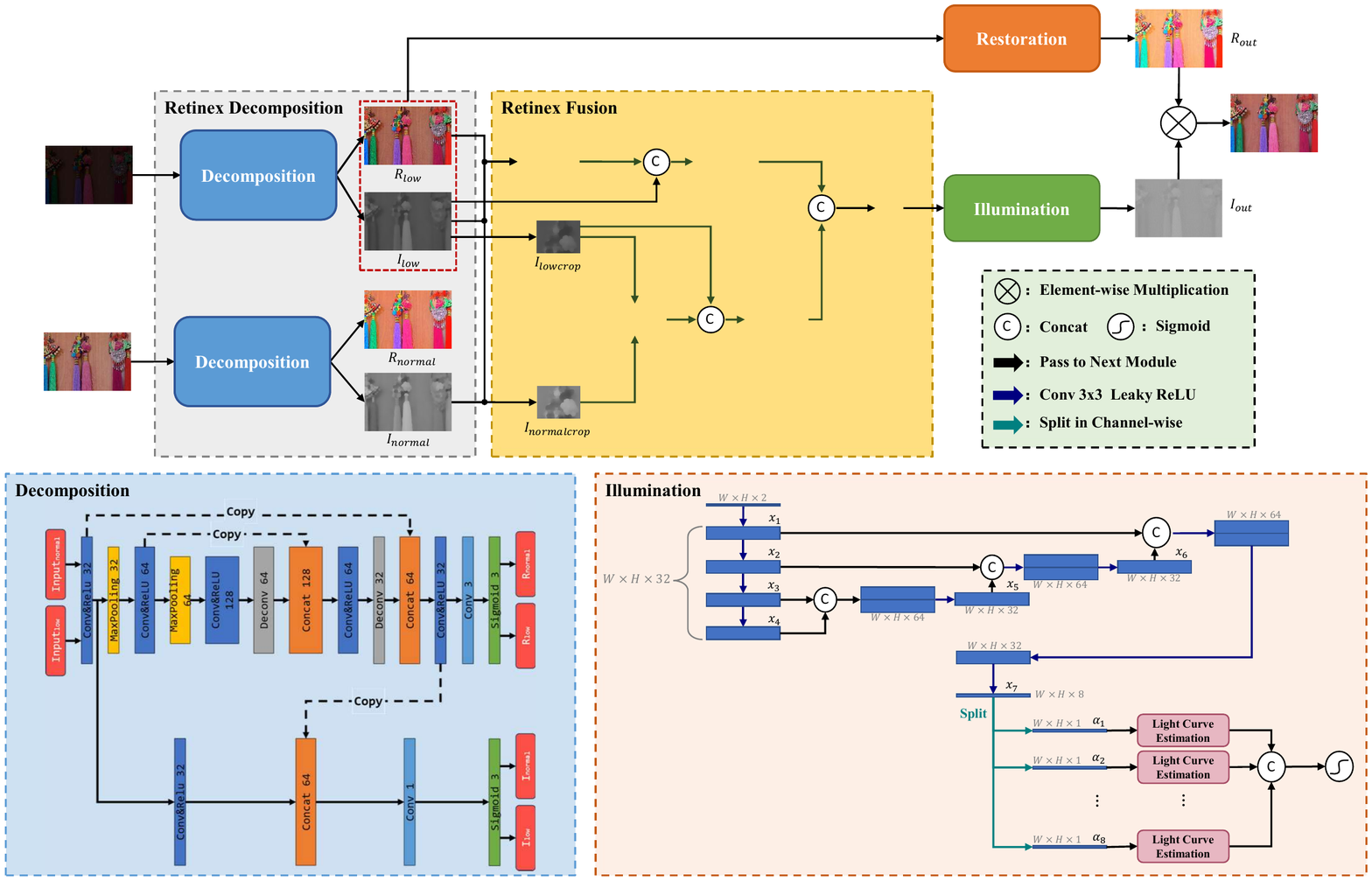}
\caption{diagram of the image fusion method}\label{fig4.2}
\end{figure}

In this paper, TV loss \cite{rudin1992nonlinear}, Light Curve estimation module, and Retinex Fusion (RF) module are applied to the KinD-Plus \cite{zhang2021beyond} network to design the KinD-LCE network. 
We reduced the noise in the output image with the total variation loss function to attach a normal term to the model, which constrains the noise by matching other loss functions. 

The Light Curve Estimation module adjusts the brightness of the illumination map, mapping different pixels to a continuously differentiable numerical space and adjusting the light curve using trainable parameters in backpropagation to suppress the chessboard effect and improve the image quality. 
The Retinex Fusion module fuses the reflection map and the illumination map in the illumination enhancement stage to compensate for the lost features. 
We have retained the original decomposition network from KinD-plus\cite{zhang2021beyond} to preliminarily extract the reflection component and illumination component from images, which are then made available for subsequent network processing and optimization. 
The restoration network is preserved to further extract the reflection component, ultimately yielding $R_{low}$. The network structure of KinD-LCE is depicted in Fig. \ref{fig4.1}.

In Fig. \ref{fig4.1}, the Retinex image decomposition network structure is shown in Retinex Decomposition and Image Fusion (RDIF) on the left side of the figure. The model is divided into three parts training the low illumination image ${\mathrm{input}}_{low}$ and the normal image ${\mathrm{input}}_{normal}$ into the Decomposition module $F_{decom}$· to obtain two sets of illumination maps $\left(I_{low}, I_{normal}\right)$ and reflection maps $\left(R_{low}, R_{normal}\right)$. The Retinex Decomposition method is shown in Fig. \ref{fig4.2}.

\begin{equation}
\begin{aligned}
I_{low},R_{low}&=F_{decom}(\text{input}_{low}), \label{eq41} \\
\end{aligned}
\end{equation}

\begin{equation}
\begin{aligned}
I_{normal},R_{normal}&=F_{decom}(\text{input}_{normal}), \label{eq42}
\end{aligned}
\end{equation}

Passing $I_{low}, R_{low}$ to Restoration module. $F_{restore}$ for restoring $R_{low}$, which is used to generate the enhanced image and also acts as a noise reduction.

\begin{equation}
\begin{aligned}
R_{out}=F_{restore}(I_{low},R_{low}), \label{eq43} \\
\end{aligned}
\end{equation}

${\text{input}}_{low},{\text{input}}_{normal}$ is passed to the RDIF module $F_{rdif}$ to restore $I_{low}$, and $x_{rf}$ is obtained by fusing $I_{low}, I_{normal}, R_{low}$ with the Retinex Fusion module, and $x_{rf}$ is fed to the Illumination module $F_{illum}$ to enhance $I_{out}$, where $F_{illum}$  uses an image enhancement method based on Light Curve Estimation, which improves the problem of noise and loss of edge detail due to the use of upsampling and downsampling.

\begin{equation}
\begin{aligned}
x_{rf}&=F_{rdif}(\text{input}t_{low}, \text{input}t_{normal}) \label{eq44} \\
\end{aligned}
\end{equation}

\begin{equation}
\begin{aligned}
I_{out}&=F_{illum}(x_{rf}), \label{eq45}
\end{aligned}
\end{equation}

Finally, based on Retinex theory, the illumination map $I_{out}$ and reflection map $R_{out}$ are fused at the pixel level to obtain the enhanced image, denoted as

\begin{equation}
\begin{aligned}
P_r=R_{out}\otimes I_{out}. \label{eq46}
\end{aligned}
\end{equation}

\subsection{TV Loss}\label{subsec42}

The TV loss is a regularization term that measures the total variation in an image. It is well-known that the total variation of a noisy image is higher than that of a noise-free image \cite{rudin1992nonlinear}. 
By minimizing the TV loss, we can achieve the goal of denoising. This is accomplished by calculating the difference between neighboring pixel values in the image. 
In this paper, we employ TV loss to reduce the noise generated during network enhancement and the noise generated during the fusion of illumination and reflection maps. 
The TV loss is defined as follows.

\begin{equation}
\begin{aligned}
L_{t\nu}=\sum_{i,j}\sqrt{\left(x_{i,j+1}-x_{i,j}\right)^2+\left(x_{i+1,j}-x_{i,j}\right)^2}. \label{eq47}
\end{aligned}
\end{equation}

In low-light images, hidden noise is prevalent, and convolutional neural networks can introduce additional noise while upsampling and downsampling the images. To mitigate this noise, we used, TV loss as it measures the total change in the image while preserving its edges. Meanwhile, Gaussian noise and Poisson noise are prevalent in the image, and Poisson noise is a kind of noise associated with illumination based on the ability to preserve the edges. The Poisson noise is illumination-related, and based on the property of preserving edges, the Gaussian noise and Poisson noise can be eliminated through TV loss. The total loss function used by the method in this paper in the Illumination module is expressed as

\begin{equation}
\begin{aligned}
L&=\beta L_{grad}+L_{mse}+L_{t\nu}, \label{eq48}\\
\end{aligned}
\end{equation}
\begin{equation}
\begin{aligned}
L_{grad}&=\left\|\left|\nabla\tilde{l}\right|-\left|\nabla I_p\right|\right\|_2^2, \label{eq49} \\
\end{aligned}
\end{equation}
\begin{equation}
\begin{aligned}
L_{mse}&=\left\|\tilde{l}-I_p\right\|_2^2, \label{eq410}
\end{aligned}
\end{equation}

Where $L_{grad}$ denotes the gradient loss, $L_{mse}$ denotes the mean square loss, $I_p$ denotes the normally illuminated image, $\widetilde{I}$ denotes the adjusted image, which is the image in the process of mapping from the low illuminated image to the normal image, $\nabla$ denotes the gradient operator, $\beta$ denotes the weight of the loss of $L_{grad}$, and the parameter $\beta=0.01$ is set in this paper according to experience.

\subsection{Retinex Fusion Module}\label{subsec43}

The Retinex Fusion module is divided into two parts to process low illumination images, local illumination coefficients $t_{local}$ are multiplied with the unit matrix $E$ to ignore spatial features and improve the effect of illumination coefficients, and global illumination coefficients $t_{global}$ are multiplied with $R_{low}$ to focus on global detail features, which work together to make the neural network focus on more image information and consider local features of the image. 
The module for image illumination attributes adjustment, which allows the edge illumination information of $R_{low}$ and $I_{low}$ at the time of combination to be complimented. It restores unsmooth areas of an image to preserve edge detail.

The global illumination matrix $S_{global}$ is first computed using $R_{low}$, and the two illumination maps $I_{low}$ and $I_{normal}$ are compared and then averaged to obtain the original illumination coefficients $t_{global}$, and then the original illumination coefficients $t_{global}$ are multiplied by $R_{low}$ to recover the missing edge detail in $I_{low}$ due to incomplete decomposition in Decomposition, $S_{global}$ is computed as follows.

\begin{equation}
\begin{aligned}
S_{global}&=R_{low}\cdot t_{global}, \label{eq411}\\
\end{aligned}
\end{equation}

\begin{equation}
\begin{aligned}
t_{global}&=\frac{1}{MN}\sum_{i,j}\frac{I_{low}(i,j)}{I_{normal}(i,j)}, \label{eq412}
\end{aligned}
\end{equation}
where $N$ and $M$ denote the height and width of $I_{low}$.

Secondly, the local illumination matrix $S_{local}$ is calculated using Eq. \eqref{eq413}, and the illumination map of the cropped low light image $\left(I_{lowcrop}\right)$ and the illumination map of the normal image $\left(I_{normalcrop}\right)$ are compared and then averaged to obtain the local illumination coefficient $t_{local}$, and then multiplied with the unit matrix $E$ of the same size as $I_{lowcrop}$ and the local illumination coefficient $t_{local}$. To cope with the detail reduction problem of $I_{low}$ under different illumination conditions and to improve the robustness, $S_{local}$ is calculated as follows.

\begin{equation}
\begin{aligned}
S_{local}&=E\cdot t_{local}, \label{eq413}\\
\end{aligned}
\end{equation}

\begin{equation}
\begin{aligned}
t_{local}&=\frac{1}{WH}\sum_{i,j}\frac{I_{lowcrop}(i,j)}{I_{nomralcrop}(i,j)}, \label{eq414}
\end{aligned}
\end{equation}

where $H$ and $W$ denote the height and width of $I_{lowcrop}$. After the above calculation, $I_{lowcrop}$ and $S_{local}$ and $I_{low}$ and $S_{global}$ are concatted to obtain $I_{flocal}$ and $I_{fglobal}$ respectively, and finally $I_{flocal}$ and $I_{fglobal}$ are concatted to obtain $x_{rf}$ and input to the light adjustment network module.

\begin{equation}
\begin{gathered}
I_{flocal}=\mathrm{Concat}(S_{local},I_{lowcrop}), \label{eq415}\\
\end{gathered}
\end{equation}

\begin{equation}
\begin{gathered}
I_{fglobal}=\mathrm{Concat}(S_{global},I_{low}), \label{eq416}\\
\end{gathered}
\end{equation}

\begin{equation}
\begin{gathered}
x_{rf}=\mathrm{Concat}(l_{flocal},l_{fglobal}), \label{eq417}
\end{gathered}
\end{equation}

The information from the reflectance map is introduced in the Retinex image fusion module to compensate for the loss of illumination map edge information because of incomplete decomposition of the decomposition network, allowing the unsmooth areas in the image to be correctly adjusted and improving the quality of the output image. The method is shown in Figure 4.2.

\subsection{Light Curve Estimation with Illumination Map}\label{subsec44}

In the study of theory, the reflectance map contains information, such as feature details and edges of the original image, and the illumination map contains information, such as contrast and brightness of the image. Therefore, further optimization of the decomposed reflection map and illumination map can achieve the effect of image enhancement. Inspired by the image editing software, the light curve was introduced to make the illumination curve of the illuminance map smoother and more natural, and each pixel point was adaptively enhanced, compared to Zero-DCE \cite{guo2020zero}, which calculates the light curve for the original image and its adjustment has an effect on the original image color, uses the Retinex theory to decompose the illumination and detail features of the image, to adjust the illumination using the light curve, and to reduce the effect of color shift.

The light curve adjustment module is different from the conventional method of Encoder-Decoder network, which needs to downsample and then upsample the image, but in the process of upsampling, it is easy to produce a "checkerboard artifacts", which causes the color of a specific part of the image to be darker than other parts of the image. 
The aforementioned situation occurs when the ratio of the convolution kernel size to the convolution step size is not an integer. 
In our work, the light curve adjustment module was used as an illumination map enhancement network to enhance the illumination map through the light curve to make the image brightness smoother and eliminate the effect of checkerboard artifacts. 
A schematic of the light curve estimation method is shown in Fig. \ref{fig4.2}.

At the beginning of the Illumination phase, the image processed by the RDIF module is used as input and finally decomposed into $8$ light estimation coefficient matrices $\alpha_i\left(i=1,2,\cdots,8\right)$ by four $3\times 3$ convolutions and a concat operation on them respectively, which partially maps the light estimation coefficient matrices into a continuous space.

The variable $\alpha_i$ in illumination model is a trainable parameter, the range belongs to $\left[-1,1\right]$, the illumination map processed by RDIF module is input to Illumination map enhancement module (Illumination), after four convolutions of $3\times 3$, the feature map obtained from each layer convolution is mapped separately, and finally the feature map with the number of channels is $x_7$, which represents the illumination estimation ratio for each pixel point on the image.

$$
\begin{aligned}
x_1&=\text{Conv}_{3\times3}(x_{rf}),\\
x_2&=\text{Conv}_{3\times3}(x_1),\\
x_3&=\text{Conv}_{3\times3}(x_2), \\
x_4&=\text{Conv}_{3\times3}(x_3),\\
x_5&=\text{Conv}_{3\times3}\big(\text{Concat}(x_3,x_4)\big), \\
x_6&=\text{Conv}_{3\times3}\big(\text{Concat}(x_2,x_5)\big),\\
x_7&=\text{Conv}_{3\times3}\big(\text{Concat}(x_1,x_6)\big),
\end{aligned}
$$

Depending on the adjustment scale factor is decomposed into $8$, which are each entered into the Light Curve Estimation module to adjust the light curve, using the formula expressed as,

\begin{equation}
\begin{aligned}
\alpha_{1},\ldots,\alpha_{8}=\mathrm{split}(x_{7}), \label{eq418}
\end{aligned}
\end{equation}
\begin{equation}
\begin{aligned}
I_i=x_{rf}(j)+\alpha_ix_{rf}(j)\left(1-x_{rf}(j)\right), \label{eq419}
\end{aligned}
\end{equation}
\begin{equation}
\begin{aligned}
I_{out}=\text{sigmoid}(\text{Concat}(I_1,[USD3P],I_8)), \label{eq420}
\end{aligned}
\end{equation}

Where $x_{rf}\left(j\right)$ represents the $j$ pixels of $x_{rf}$ , $I_{out}$ represents the adjusted image, and $I_i$ represents the image adjusted by the $i$ th $\alpha_i$.

\section{Experimental Results and Discussion}\label{sec5}

\subsection{Implemental Details}\label{subsec51}

\textbf{Datasets \& Settings.} 
We conduct experiments on two datasets: LOw-Light dataset \cite{wei2018deep} and GladNet \cite{wang2004image} dataset. The GladNet dataset, proposed by Wang et al. \cite{wang2004image}, contains $5000$ pairs of 8-bit RGB image pairs of low-light/normal images, which we use as the training data. We split the dataset into $3500$ pairs for training, $1000$ pairs for validation, and $500$ pairs for testing. 
The LOw-Light dataset \cite{wei2018deep}, which contains $500$ pairs of low light/normal image pairs, is used as the comparison set for the test set to validate the enhancement effect. All images in both datasets are captured from real scenes. 
The GladNet dataset has variable image sizes with various scales, while all the images in the LOw-Light dataset are $600 \times 400 \times 3$ images. 
For the downstream task, we use semantic segmentation on the ExDark \cite{loh2019getting} dataset, which is commonly used for blackout scenes. 
This dataset includes images of black night scenes taken from $13$ regions, with over $10$ categories of scenes such as buildings, streets, and cars.

\textbf{Metrics.} Peak signal-to-noise ratio (PSNR) \cite{huynh2008scope}, structural similarity index (SSIM) \cite{wang2004image}, mean absolute error (MAE) and mean squared error (MSE) are widely used evaluation metrics in image processing and computer vision tasks.
PSNR is a measure of the peak error between two images and is often used to evaluate the image reconstruction quality, where higher values indicate better image quality. 
SSIM evaluates the similarity between two images by comparing the luminance, contrast, and structure of the images. 
MAE and MSE measure the difference between two images by computing the absolute and squared differences between the pixel values, respectively. 
These metrics are commonly used to evaluate the accuracy of image restoration and enhancement methods, where lower values indicate better performance.

\textbf{Experimental Environment.} 
In this section, all experimental data except for the data generated under Nvidia NX for model optimization deployment testing, were generated under Geforce RTX 2080ti. Some of the data may differ from the data in the original paper.

\subsection{Ablation Study}\label{subsec52}

This section discusses our ablation experiments on the GladNet and LOw-Light datasets to validate the proposed improvements. We then evaluated the model using the validation sets of both datasets.

\begin{figure}[h]%
\centering
\includegraphics[width=0.9\textwidth]{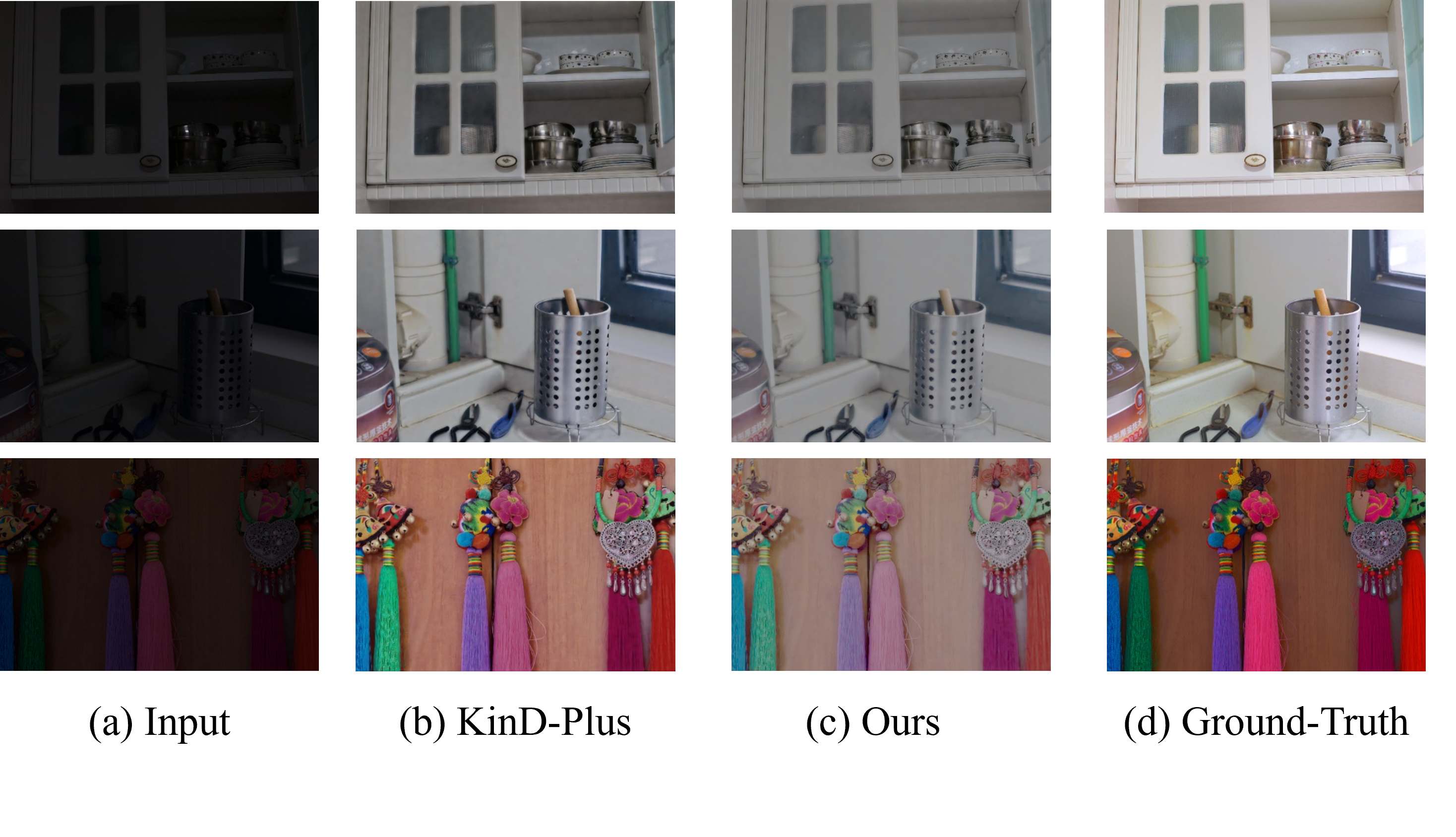}
\caption{The effect of the light curve estimation module: Figure (a) represents the image before processing, all under low light scenes; figure (b) shows the image after KinD-Plus \cite{zhang2021beyond} algorithm; Figure (c) shows the enhancement effect of our method, after using the light curve estimation; Figure (d) shows the normal image corresponding to Figure (a), i.e., Ground Truth. adding the light curve estimation module to the Illumination module, from Figure (c), we can see that adding the light curve estimation module enhances the original darker shadows, for the overall brightness of the image is significantly improved and the effect is better than Fig. (b).}\label{fig5.1}
\end{figure}

\textbf{Ablation for Light Curve Estimation (LCE).} In studying the KinD-Plus \cite{zhang2021beyond} network, we found that the illumination map enhancement module did not consider the color and light balance, so we designed its illumination map enhancement network with light curve estimation and our proposed KinD-LCE network. Our LCE module improved the PSNR metrics for the same image input. The rendering is shown in Fig. \ref{fig5.1}, and the metrics table is shown in Tab. \ref{tab5.1}.

\begin{table}[h]
\caption{PNSR and SSIM metrics of using Light Curve Estimation}\label{tab5.1}%
\begin{tabular}{@{}lll@{}}
\toprule
Algorithm & PNSR ↑  & SSIM ↑ \\
\midrule
KinD-Plus w/o LCE    & 16.2156   & 0.8173   \\
KinD-plus w/ LCE    &  18.9548   & 0.8112  \\
\botrule
\end{tabular}
\end{table}

\textbf{Ablation for Retinex Fusion.} We fused the reflectance map with the illumination map, adjusted the coefficients of illumination map enhancement with the pixel values of the reflectance map, and compensated for the loss of edge details and other problems generated from the original image decomposition by fusing the reflectance and illumination maps. The image fusion module improved the PSNR and SSIM metrics over the original KinD-Plus \cite{zhang2021beyond} network by $16.66\%$ and $3.52\%$, respectively. The results are shown in Fig. \ref{fig5.1}, and the metrics table is shown in Tab. \ref{tab5.1}.

\begin{table}[h]
\caption{PNSR and SSIM metrics of using Retinex Fusion}\label{tab5.2}%
\begin{tabular}{@{}lll@{}}
\toprule
Algorithm & PNSR ↑  & SSIM ↑ \\
\midrule
KinD-Plus w/o Fusion    & 16.2156   & 0.8173   \\
KinD-plus w/ Fusion    &  19.4154   & 0.8433  \\
\botrule
\end{tabular}
\end{table}

\begin{figure}[h]%
\centering
\includegraphics[width=0.9\textwidth]{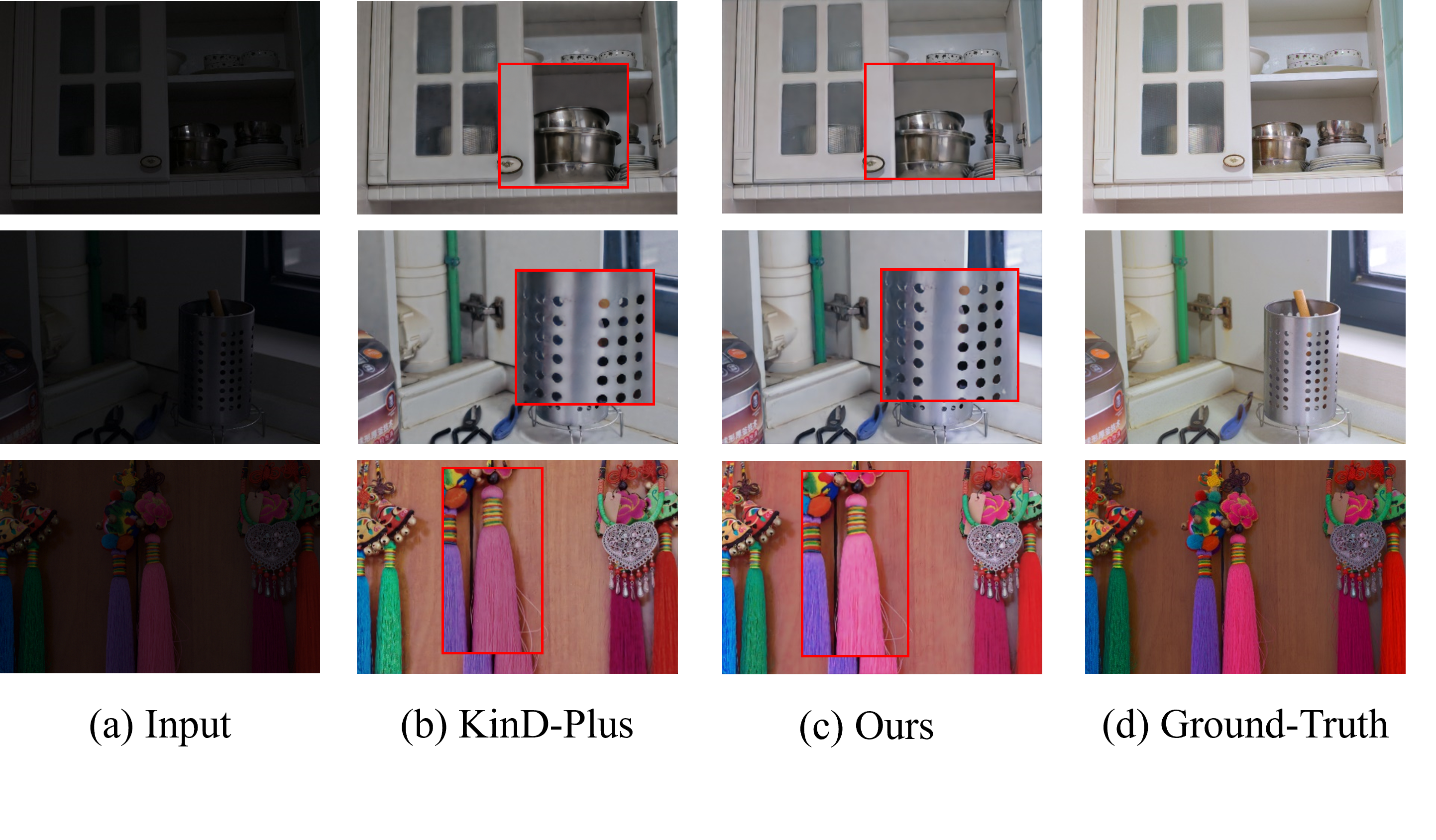}
\caption{shows the effect of the Retinex Fusion module. In panel (a), we see the original low-light images before processing. In panel (b), the images are enhanced using the KinD-Plus \cite{zhang2021beyond} algorithm. In panel (c), we show the results of our proposed approach, which includes the addition of the Retinex Fusion (RF) module to the network. Panel (d) shows the corresponding ground truth images. We observe that the darker areas of the image are successfully enhanced after image fusion in panel (c). Compared with the results in panel (b), the image enhancement is more effective, and the details in the darker areas are clearer and closer to those of the ground truth images in panel (d).}\label{fig5.2}
\end{figure}

\textbf{Ablation for TV loss.} We add TV loss to the Illumination module to calculate the total variation of the illumination map to denoise the illumination map and reduce the noise due to the decomposition network, the metrics table is shown in Tab. \ref{tab5.3}.

\begin{table}[h]
\caption{PNSR and SSIM metrics of using TV loss}\label{tab5.3}%
\begin{tabular}{@{}llll@{}}
\toprule
Algorithm & PNSR ↑  & SSIM ↑ \\
\midrule
KinD-Plus w/o TV Loss    & 16.2156   & 0.8173   \\
KinD-plus w/ TV Loss    &  18.5241   & 0.7822  \\
\botrule
\end{tabular}
\end{table}

\begin{figure}[h]%
\centering
\includegraphics[width=0.9\textwidth]{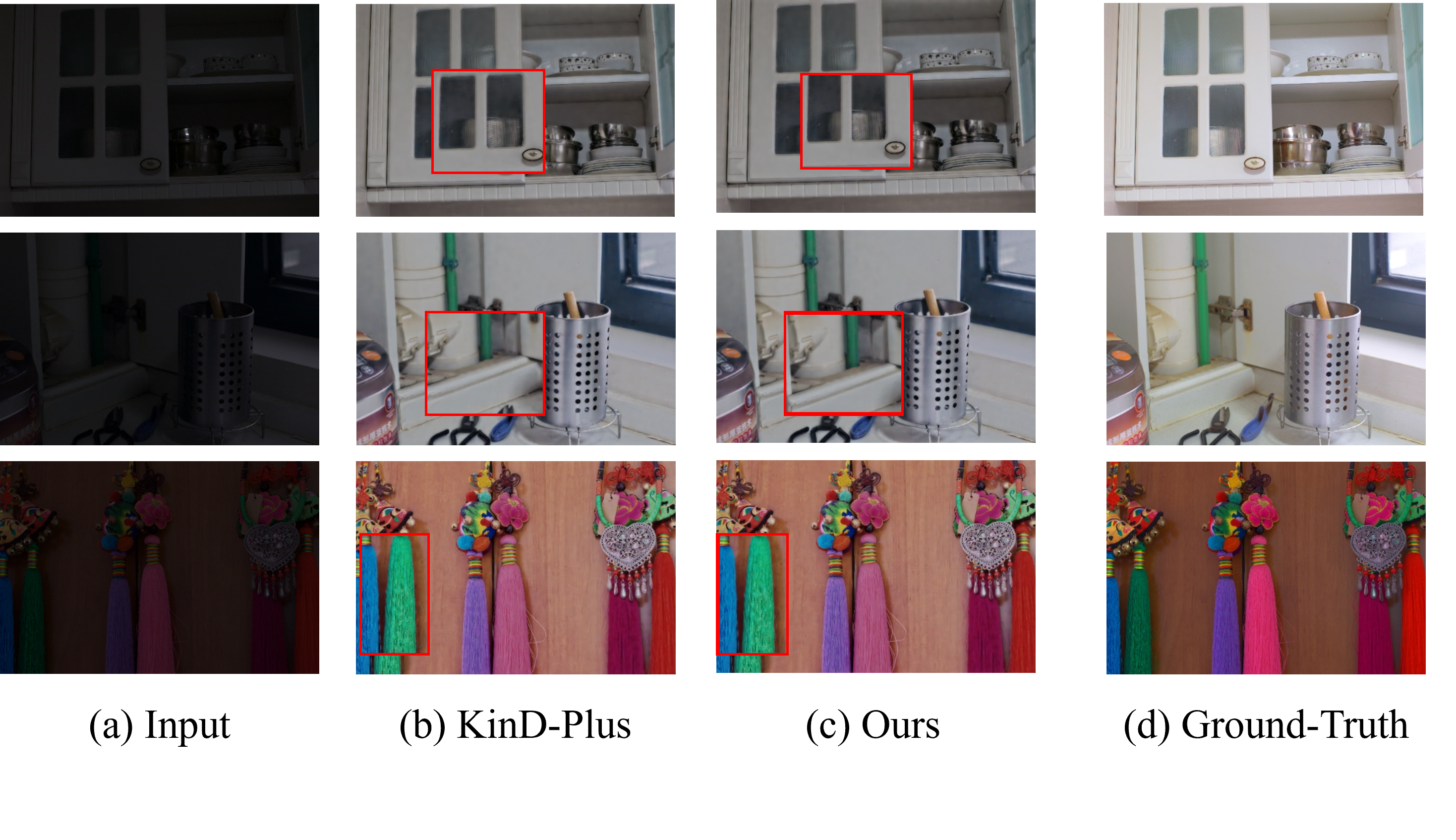}
\caption{Presents a comparison of the results obtained using KinD-LCE and KinD-Plus [14] algorithms on low-light scenes. Fig. (a) shows the original images before processing, while Fig. (b) shows the images after processing using KinD-Plus. Our proposed method is illustrated in Fig. (c), and the ground truth image corresponding to Fig. (a) is shown in Fig. (d). From Fig. (c), we observe that the proposed method produces more detailed enhancements compared to KinD-Plus (Fig. (b)) while exhibiting lower levels of noise and higher brightness.}\label{fig5.3}
\end{figure}

\subsection{Comparison of KinD-LCE and Other Methods}\label{subsec53}

After our experimental comparison and cross-validation using different methods, we can see that we can achieve relatively high metrics by adding the three modules proposed in this paper. The experimental results are shown in Tab. \ref{tab5.4}.

\begin{table}[h]
\caption{PNSR and SSIM metrics of using different methods}\label{tab5.4}%
\begin{tabular}{@{}lllll@{}}
\toprule
Light Curve Estimation & Fusion & TV Loss & PNSR ↑  & SSIM ↑ \\
\midrule
\checkmark &  &  & 16.2156 & 0.8173 \\
 & \checkmark &  & 18.9548 & 0.8112 \\
 &  & \checkmark & 19.4154 & 0.8433 \\
 \checkmark & \checkmark &  & 18.2518 & 0.7912 \\
 & \checkmark & \checkmark & 17.4779 & 0.8113 \\ 
 \checkmark & \checkmark & \checkmark & 19.7216 & 0.8213 \\ 
\botrule
\end{tabular}
\end{table}

The method described in this paper was compared with the best methods available, the enhancement effect was evaluated using PSNR and SSIM metrics, and the evaluation was tested using 15 validation sets of the Low-Light dataset. The images are shown in Figure \ref{tab5.3}, and the metrics are shown in Table \ref{tab5.5}. According to Tab. \ref{tab5.5}, we were performing at a high level in terms of metrics, providing evidence for the enhancement effect of our method on low-light tasks. This evidence is consistent with our observations, but we must improve the robustness of our method. Although there has been some optimization, some regions in some images HAVE color and detail variations.

\begin{table}[h]
\caption{Table of comparative indicators for other methods, with "↑" indicating larger is better and "↓" indicating smaller is better.}\label{tab5.5}%
\begin{tabular}{@{}lllll@{}}
\toprule
Algorithm & PNSR ↑  & SSIM ↑ & MAE ↓ & MSE ↓\\
\midrule
KinD-plus\cite{zhang2021beyond}    & 16.2156   & 0.8023 & 101.3138 & 103.6879  \\
MBLLEN \cite{lv2018mbllen}    &  17.8583   & 0.7247 & 97.1637 & \color{red}{\textbf{77.0107}} \\
RetinexNet \cite{liu2021retinex}  &  14.9774   & 0.5392 & 120.9511 & 102.1184 \\
EnglightenGAN \cite{jiang2021enlightengan}    &  18.9248   & 0.8112 & 147.7995 & 96.0031 \\
Zero-DCE \cite{guo2020zero}    &  14.7971   & 0.6640 & \color{red}{\textbf{68.2873}} & 108.8530 \\
Ours    &  \color{red}{\textbf{19.7216}}   & \color{red}{\textbf{0.8213}} & 101.1897 & 108.1573 \\
\botrule
\end{tabular}
\end{table}

\subsection{Comparison in Downstream Task}\label{subsec54}

Applying the method of this paper to downstream tasks (e.g., object detection and semantic segmentation) for conventional dark images without image processing can significantly reduce the detection effectiveness of our method. Therefore, using our method, the dark image was processed and then applied to the downstream task regardless of whether an object was detected in the image.

To highlight the effectiveness of the algorithms in this paper, we experimentally compared the images using different low-light enhancement networks to verify the effectiveness of various low-light enhancement networks for downstream tasks. The downstream task validation experiment used the HRNet \cite{yuan2020object} network for semantic segmentation with low-light images in traveling scenes for enhancement. We used the ExDark \cite{loh2019getting} dataset for this experiment when using the Cityscape dataset-based \cite{cordts2016cityscapes}. The trained model was validated against the driving scene. The following are the experimental comparison plots.

\begin{figure}[h]%
\centering
\includegraphics[width=0.9\textwidth]{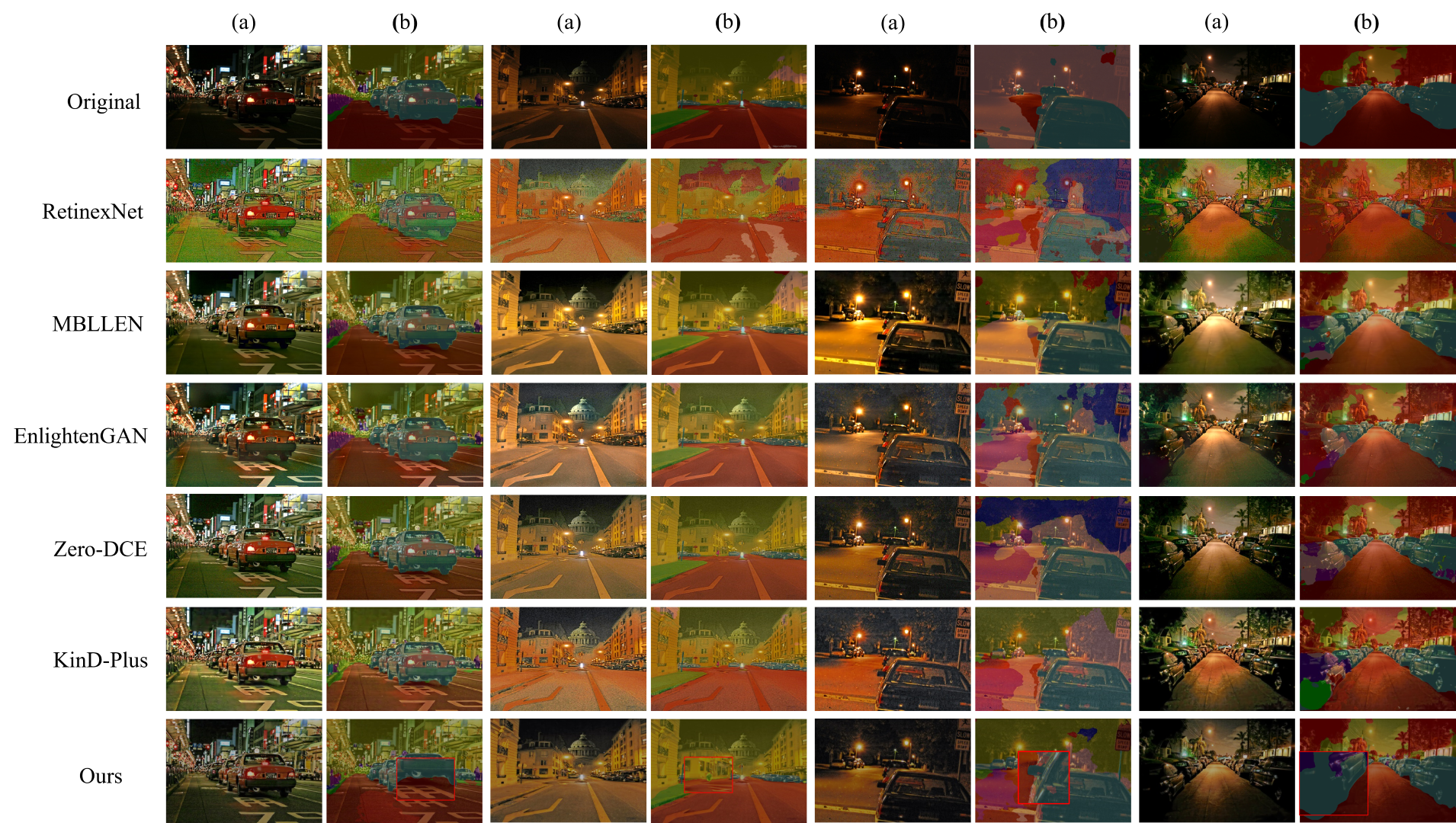}
\caption{presents a comparison of the downstream experiment results. Specifically, the images in column (a) show the output of the corresponding network after low-light image enhancement, while the images in column (b) represent the images after semantic segmentation using HRNet. Our proposed method achieves the best performance in terms of semantic segmentation, producing images with clearer segmented edge contours and more accurate classified categories.}\label{fig5.4}
\end{figure}

Based on the implementation results (Fig. \ref{fig5.4}), our algorithm performed well in color restoration and edge information restoration in downstream tasks, making our algorithm more effective for practical downstream applications. However, it is undeniable that in some extreme cases, many noises and brightness differences in the original picture information cause our algorithm to produce color distortion. We believe that the optimization of the fusion process could more effectively solve these problems.

\subsection{Test in Model Optimization Deployment}\label{subsec55}

The method we tested for our paper's model optimization deployment can verify the superiority of our method when deployed on edge devices. We chose Nvidia NX as our test device and the Low-Light dataset as our test data. We deployed our code on it and ran it, ultimately obtaining the following results, as shown in Tab.\ref{tab5.6}:

\begin{table}[h]
\caption{The performance of our method before and after model optimization }\label{tab5.6}%
\begin{tabular}{@{}lll@{}}
\toprule
 & Before Optimization  & Optimized \\
\midrule
FPS    & 0.306   & 23   \\
\botrule
\end{tabular}
\end{table}

\begin{figure}
    \centering
    \includegraphics[width=0.75\textwidth]{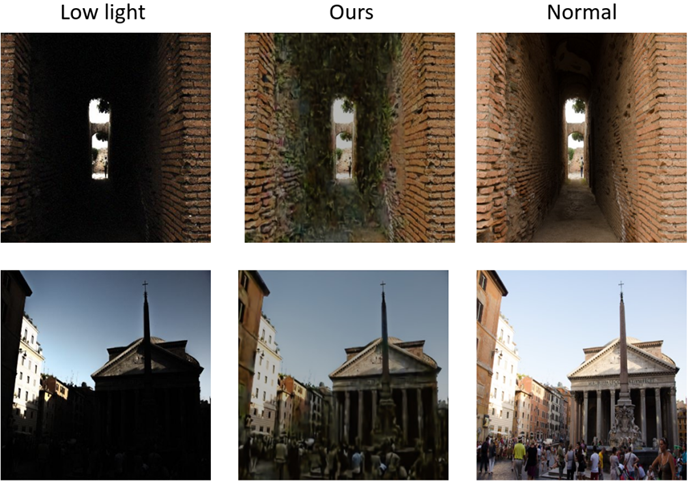}
    \caption{Results of illustrations under challenging low-light conditions.}
    \label{fig5.5}
\end{figure}

After deploying and optimizing the method described in this paper on edge devices, its average frame rate performance has significantly improved, indicating its potential in edge computing tasks.

\subsection{Discussion}\label{subsec56}
When encountering significant color information loss, such as large areas of featureless black pixel blocks, as illustrated in Fig. \ref{fig5.5}, our proposed method exhibits notable color distortion. 
We acknowledge that our method still has certain limitations in global modeling, which prevents us from accurately restoring invalid information in images when the existing valid information is scarce or of poor quality.

\section{Conclusions}\label{sec6}

This paper proposes the KinD-LCE network, which integrates light curve evaluation, light map, and reflectance map fusion for low-light enhancement tasks. Our approach improved upon the KinD-Plus \cite{zhang2021beyond} method by compensating for the introduced noise and achieving effective image illumination enhancement. 
Specifically, we adjusted the light map using light curve estimation and fused the features of the reflectance map into the light map to improve the brightness of the image while reducing noise. 
We demonstrated the effectiveness of our method through ablation experiments and experiments on downstream tasks. However, the performance of our method has limitations when evaluated with metrics such as MAE and MSE, as image quality can still be compromised during image processing. 
Due to the limitation of the decomposition network, it is unable to achieve complete decomposition, while Retinex theory requires that the image be completely separated into illumination and reflection components. Even though we have implemented optimization measures in this regard, subjectively, this issue still inevitably occurs. 
Besides, in the future, this research will try to address this problem and optimize possible limitations in global modeling.

\section*{Declarations}

\subsection*{Ethical Approval}
this declaration is “not applicable”. 

\subsection*{Competing interests}
No potential conflict of interest was reported by the authors

\subsection*{Authors’ Contributions}
Conceptualization, Xiaochun Lei. and Junlin Xie.; Data Curation, Junlin Xie. and Weiliang Mai; Formal Analysis, Xiaochun Lei. and Junlin Xie.; Methodology, Junlin Xie. and Xiaochun Lei.; Supervision, Zetao Jiang. and Zhaoting Gong.; Validation, Ziqi Shan. and Linjun Lu.; Visualization, Linjun Lu. and Chang Lu, Zhaoting Gong. and Weiliang Mai.; Writing original draft preparation, Xiaochun Lei. and Junlin Xie.; Writing review and editing, Zetao Jiang., Zhaoting Gong. and Weiliang Mai; All authors have read and agreed to the published version of the manuscript.

\subsection*{Funding}

This work was supported by the National Natural Science Foundation of China (61876049, 62172118) and Nature Science Key Foundation of Guangxi (2021GXNSFDA196002); in part by the Sichuan Regional Innovation Cooperation Project (2021YFQ0002); in part by the Guangxi Key Laboratory of Image and Graphic Intelligent Processing under Grants (GIIP2004) and Student’s Platform for Innovation and Entrepreneurship Training Program under Grant (202010595053, 202010595168, S202110595168, 202110595025, 202310595036).

\subsection*{Availability of data and materials}

The data that support the findings of this study are available on request from the corresponding author, Jiang, upon reasonable request.

\bibliographystyle{ieeetr}
\bibliography{./ref} 

\begin{thebibliography}{10}

\bibitem{pizer1987adaptive}
S.~M. Pizer, E.~P. Amburn, J.~D. Austin, R.~Cromartie, A.~Geselowitz, T.~Greer, B.~ter Haar~Romeny, J.~B. Zimmerman, and K.~Zuiderveld, ``Adaptive histogram equalization and its variations,'' {\em Computer vision, graphics, and image processing}, vol.~39, no.~3, pp.~355--368, 1987.

\bibitem{land1971lightness}
E.~H. Land and J.~J. McCann, ``Lightness and retinex theory,'' {\em Josa}, vol.~61, no.~1, pp.~1--11, 1971.

\bibitem{zhou2020multiscale}
J.~Zhou, D.~Zhang, and W.~Zhang, ``Multiscale fusion method for the enhancement of low-light underwater images,'' {\em Mathematical Problems in Engineering}, vol.~2020, pp.~1--15, 2020.

\bibitem{jobson1997multiscale}
D.~J. Jobson, Z.-u. Rahman, and G.~A. Woodell, ``A multiscale retinex for bridging the gap between color images and the human observation of scenes,'' {\em IEEE Transactions on Image processing}, vol.~6, no.~7, pp.~965--976, 1997.

\bibitem{guo2020zero}
C.~Guo, C.~Li, J.~Guo, C.~C. Loy, J.~Hou, S.~Kwong, and R.~Cong, ``Zero-reference deep curve estimation for low-light image enhancement,'' in {\em Proceedings of the IEEE/CVF conference on computer vision and pattern recognition}, pp.~1780--1789, 2020.

\bibitem{rahman2020efficient}
Z.~Rahman, P.~Yi-Fei, M.~Aamir, S.~Wali, and Y.~Guan, ``Efficient image enhancement model for correcting uneven illumination images,'' {\em IEEE Access}, vol.~8, pp.~109038--109053, 2020.

\bibitem{li2021learning}
C.~Li, C.~Guo, and C.~C. Loy, ``Learning to enhance low-light image via zero-reference deep curve estimation,'' {\em IEEE Transactions on Pattern Analysis and Machine Intelligence}, vol.~44, no.~8, pp.~4225--4238, 2021.

\bibitem{zhang2019kindling}
Y.~Zhang, J.~Zhang, and X.~Guo, ``Kindling the darkness: A practical low-light image enhancer,'' in {\em Proceedings of the 27th ACM international conference on multimedia}, pp.~1632--1640, 2019.

\bibitem{zhang2021beyond}
Y.~Zhang, X.~Guo, J.~Ma, W.~Liu, and J.~Zhang, ``Beyond brightening low-light images,'' {\em International Journal of Computer Vision}, vol.~129, pp.~1013--1037, 2021.

\bibitem{jiang2021enlightengan}
Y.~Jiang, X.~Gong, D.~Liu, Y.~Cheng, C.~Fang, X.~Shen, J.~Yang, P.~Zhou, and Z.~Wang, ``Enlightengan: Deep light enhancement without paired supervision,'' {\em IEEE transactions on image processing}, vol.~30, pp.~2340--2349, 2021.

\bibitem{liu2021retinex}
R.~Liu, L.~Ma, J.~Zhang, X.~Fan, and Z.~Luo, ``Retinex-inspired unrolling with cooperative prior architecture search for low-light image enhancement,'' in {\em Proceedings of the IEEE/CVF Conference on Computer Vision and Pattern Recognition}, pp.~10561--10570, 2021.

\bibitem{xie2020semantically}
J.~Xie, H.~Bian, Y.~Wu, Y.~Zhao, L.~Shan, and S.~Hao, ``Semantically-guided low-light image enhancement,'' {\em Pattern Recognition Letters}, vol.~138, pp.~308--314, 2020.

\bibitem{rahman2021structure}
Z.~Rahman, Y.-F. Pu, M.~Aamir, and S.~Wali, ``Structure revealing of low-light images using wavelet transform based on fractional-order denoising and multiscale decomposition,'' {\em The Visual Computer}, vol.~37, no.~5, pp.~865--880, 2021.

\bibitem{guan2021dual}
Y.~GUAN, M.~Aamir, Z.~Rahman, Z.~A. Dayo, W.~A. Abro, M.~Ishfaq, and Z.~HU, ``A dual-tree complex wavelet transform-based model for low-illumination image enhancement,'' {\em Wuhan University Journal of Natural Sciences}, vol.~26, no.~05, pp.~405--414, 2021.

\bibitem{liu2016image}
Y.~Liu, X.~Chen, R.~K. Ward, and Z.~J. Wang, ``Image fusion with convolutional sparse representation,'' {\em IEEE signal processing letters}, vol.~23, no.~12, pp.~1882--1886, 2016.

\bibitem{yan2016network}
M.~Yan, C.~A. Chan, W.~Li, I.~Chih-Lin, S.~Bian, A.~F. Gygax, C.~Leckie, K.~Hinton, E.~Wong, and A.~Nirmalathas, ``Network energy consumption assessment of conventional mobile services and over-the-top instant messaging applications,'' {\em IEEE Journal on Selected Areas in Communications}, vol.~34, no.~12, pp.~3168--3180, 2016.

\bibitem{yan2019pecs}
M.~Yan, W.~Li, C.~A. Chan, S.~Bian, I.~Chih-Lin, and A.~F. Gygax, ``Pecs: Towards personalized edge caching for future service-centric networks,'' {\em China Communications}, vol.~16, no.~8, pp.~93--106, 2019.

\bibitem{ying2017bio}
Z.~Ying, G.~Li, and W.~Gao, ``A bio-inspired multi-exposure fusion framework for low-light image enhancement,'' {\em arXiv preprint arXiv:1711.00591}, 2017.

\bibitem{ying2017new}
Z.~Ying, G.~Li, Y.~Ren, R.~Wang, and W.~Wang, ``A new image contrast enhancement algorithm using exposure fusion framework,'' in {\em Computer Analysis of Images and Patterns: 17th International Conference, CAIP 2017, Ystad, Sweden, August 22-24, 2017, Proceedings, Part II 17}, pp.~36--46, Springer, 2017.

\bibitem{gharbi2017deep}
M.~Gharbi, J.~Chen, J.~T. Barron, S.~W. Hasinoff, and F.~Durand, ``Deep bilateral learning for real-time image enhancement,'' {\em ACM Transactions on Graphics (TOG)}, vol.~36, no.~4, pp.~1--12, 2017.

\bibitem{li2021low}
C.~Li, C.~Guo, L.~Han, J.~Jiang, M.-M. Cheng, J.~Gu, and C.~C. Loy, ``Low-light image and video enhancement using deep learning: A survey,'' {\em IEEE transactions on pattern analysis and machine intelligence}, vol.~44, no.~12, pp.~9396--9416, 2021.

\bibitem{guo2016lime}
X.~Guo, Y.~Li, and H.~Ling, ``Lime: Low-light image enhancement via illumination map estimation,'' {\em IEEE Transactions on image processing}, vol.~26, no.~2, pp.~982--993, 2016.

\bibitem{lore2017llnet}
K.~G. Lore, A.~Akintayo, and S.~Sarkar, ``Llnet: A deep autoencoder approach to natural low-light image enhancement,'' {\em Pattern Recognition}, vol.~61, pp.~650--662, 2017.

\bibitem{rudin1992nonlinear}
L.~I. Rudin, S.~Osher, and E.~Fatemi, ``Nonlinear total variation based noise removal algorithms,'' {\em Physica D: nonlinear phenomena}, vol.~60, no.~1-4, pp.~259--268, 1992.

\bibitem{wei2018deep}
C.~Wei, W.~Wang, W.~Yang, and J.~Liu, ``Deep retinex decomposition for low-light enhancement,'' {\em arXiv preprint arXiv:1808.04560}, 2018.

\bibitem{wang2004image}
Z.~Wang, A.~C. Bovik, H.~R. Sheikh, and E.~P. Simoncelli, ``Image quality assessment: from error visibility to structural similarity,'' {\em IEEE transactions on image processing}, vol.~13, no.~4, pp.~600--612, 2004.

\bibitem{loh2019getting}
Y.~P. Loh and C.~S. Chan, ``Getting to know low-light images with the exclusively dark dataset,'' {\em Computer Vision and Image Understanding}, vol.~178, pp.~30--42, 2019.

\bibitem{huynh2008scope}
Q.~Huynh-Thu and M.~Ghanbari, ``Scope of validity of psnr in image/video quality assessment,'' {\em Electronics letters}, vol.~44, no.~13, pp.~800--801, 2008.

\bibitem{lv2018mbllen}
F.~Lv, F.~Lu, J.~Wu, and C.~Lim, ``Mbllen: Low-light image/video enhancement using cnns.,'' in {\em BMVC}, vol.~220, p.~4, 2018.

\bibitem{yuan2020object}
Y.~Yuan, X.~Chen, and J.~Wang, ``Object-contextual representations for semantic segmentation,'' in {\em Computer Vision--ECCV 2020: 16th European Conference, Glasgow, UK, August 23--28, 2020, Proceedings, Part VI 16}, pp.~173--190, Springer, 2020.

\bibitem{cordts2016cityscapes}
M.~Cordts, M.~Omran, S.~Ramos, T.~Rehfeld, M.~Enzweiler, R.~Benenson, U.~Franke, S.~Roth, and B.~Schiele, ``The cityscapes dataset for semantic urban scene understanding,'' in {\em Proceedings of the IEEE conference on computer vision and pattern recognition}, pp.~3213--3223, 2016.

\end{thebibliography}

\end{document}